\documentclass[letterpaper, 10 pt, conference]{ieeeconf}
\IEEEoverridecommandlockouts
\let\labelindent\relax
\usepackage{enumitem}
\usepackage{balance}
\usepackage[caption=false,font=footnotesize]{subfig}
\usepackage{array}
\usepackage{textcomp}
\usepackage{mathtools, nccmath}
\usepackage{graphicx}
\usepackage{amsfonts}
\usepackage{autobreak}
\usepackage{amsmath,amssymb,bm}
\usepackage{tikz}
\usepackage{arydshln}
\usepackage{multirow}
\usepackage{epstopdf}
\usepackage{cite}
\usepackage{url}
\usepackage{siunitx}
\usepackage{xcolor}
\usepackage{float}
\usepackage{stfloats}
\usepackage{colortbl}
\usepackage{booktabs}
\usepackage{algorithm}
\usepackage{algpseudocode}

\algrenewcommand\alglinenumber[1]{\footnotesize #1.}
\makeatletter
\let\NAT@parse\undefined
\makeatother
\usepackage[hidelinks]{hyperref}

\allowdisplaybreaks[4]

\title{\LARGE \bf Language-Guided Terrain-Adaptive Neural MPC for Autonomous Traversal of Articulated Tracked Robots}

\author{Zhenfeng Gan$^{*}$, Yanbo Chen$^{*}$, Lirong Che, Yongyi Ma, Rongkai Zhu, and Xueqian Wang$^{\dagger}$%
}

\begin{document}

\maketitle
\begin{abstract}
In urban search and rescue, articulated tracked robots (ATRs) must traverse structured but contact-rich environments such as stairwells and cluttered building interiors. Reliable autonomy remains challenging because robot--terrain interaction (RTI) is hybrid and discontinuous, and effective flipper--track coordination is difficult to model analytically.
We present a language-guided terrain-adaptive neural model predictive control (MPC) framework for autonomous traversal. A terrain-conditioned neural kinematics model predicts short-horizon task-state increments from a local height sequence and recent trajectories; neural MPC plans with multi-objective costs and strict feasibility constraints; and a large language model (LLM) enables terrain adaptation by proposing bounded updates to selected weights and bounds through a safety-checked interface with range clipping, rate limiting, and consistency checks. The compiled predictor enables a full control cycle within 100\,ms.
Across three traversal tasks and a multi-height generalization setting, the proposed terrain-adaptive neural MPC improves an aggregate traversal-quality score by up to 71\% over non-adaptive neural MPC and by 67\% over a PPO baseline, while eliminating measurable collision impacts during descent. These results indicate that combining terrain-conditioned neural kinematics, optimization-based planning, and language-guided adaptation yields data-efficient and robust autonomy for articulated tracked robots.
Real-robot trials over four indoor obstacles further demonstrate transfer to contact-rich physical traversal.
\end{abstract}


\section{Introduction}

Recent vision-language-action (VLA) models show that large-scale visual and linguistic priors can be transferred to robotic control by mapping observations and language instructions directly to actions~\cite{Brohan2023RT2,Kim2024OpenVLA}. This paradigm is promising for semantic generalization, but it is not sufficient by itself for contact-rich autonomy. In many deployments, the action generator is executed as a feedforward policy conditioned on the current observation; it does not explicitly optimize over future contact events, actuator limits, body-attitude bounds, or task-specific safety constraints. Consequently, when the robot encounters a new terrain geometry or a changed traversal preference, the observation can become out of distribution and the policy may produce actions that are semantically plausible but dynamically infeasible.

\begin{figure}[t]
    \centering
    \includegraphics[width=.9\linewidth]{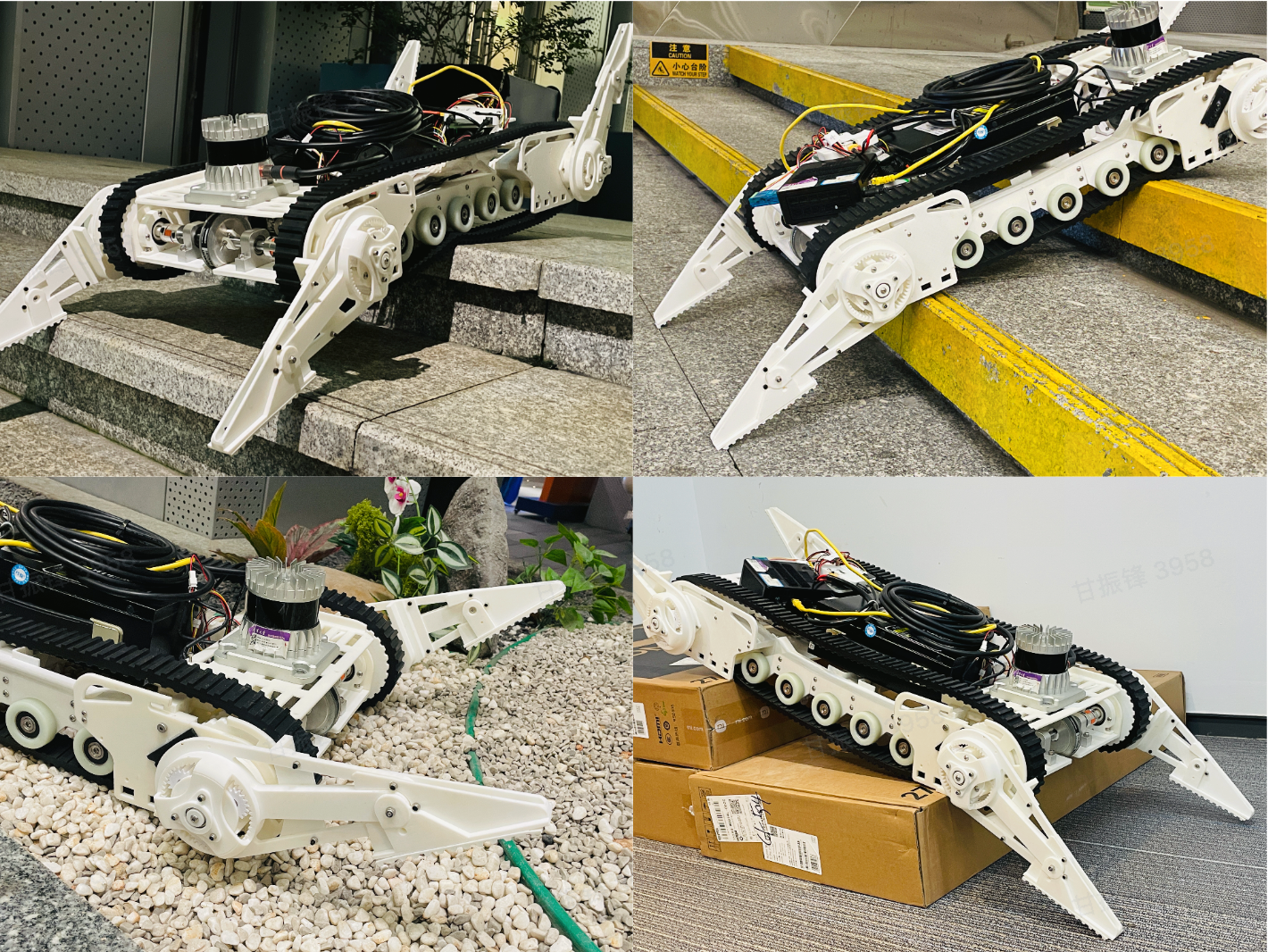}
    \caption{Articulated tracked robot studied in this work.}
    \label{fig:robot_overview}
\end{figure}

This limitation is especially pronounced in urban search and rescue, where articulated tracked robots (ATRs) must traverse stairwells, landings, and cluttered building interiors~\cite{Klamt2019Centauro,Michael2012Collaborative,Niroui2019DeepRL}. Their articulated flippers enlarge the contact patch, improve traction, and facilitate obstacle traversal~\cite{Delmerico2019Rescue,Ugenti2023AllTerrain} (Fig.~\ref{fig:robot_overview}), but reliable motion requires coordinated chassis--flipper behavior under hybrid robot--terrain interaction (RTI). Frequent support-mode switches and discontinuous contacts make explicit modeling difficult, while small geometric variations can change the desired trade-off between stability, impact, and speed. These factors motivate a controller that can retain hard feasibility constraints while adapting its objective to different environments and operator preferences.

Classical geometry-based planners ensure basic traversability but rely on pre-defined robot configurations, which limits optimality and often produces non-smooth trajectories on irregular terrain~\cite{Okada2009IROS,Okada2011JFR,Gianni2016JFR}. Reinforcement learning avoids explicit modeling yet commonly exhibits jerky switching in discrete action spaces, oscillations in continuous action spaces, and requires terrain-specific reward shaping~\cite{10333976,Mitriakov2021Staircase,articlerl,Pecka2016IROS,Pecka2018RAL,Nguyen2019RLReview,Kalashnikov2018QTOpt}. Vision-language policies improve semantic grounding, but direct action prediction still lacks an explicit mechanism for enforcing traversal constraints and retuning objective trade-offs under out-of-distribution terrain.

Model predictive control (MPC) offers a complementary route: it enforces hard constraints, coordinates multiple objectives via tunable cost weights, and yields smooth behavior through receding-horizon optimization. However, practical MPC deployment on ATRs faces two obstacles. First, accurate modeling under hybrid contacts is difficult; non-smooth contact transitions and support changes make real-time numerical optimization challenging. Second, tuning is labor-intensive: weights and constraints in a multi-objective formulation often require scenario-specific adjustment by experts. To address these challenges, we present a \textbf{language-guided terrain-adaptive neural MPC} framework for ATRs. Prior work has shown that neural NMPC can be deployed in safety-critical domains such as automated driving under unknown friction conditions~\cite{9638389}. We augment MPC with a terrain-conditioned neural kinematic predictor to bypass explicit contact modeling and introduce a regulation layer powered by a large language model (LLM). Rather than generating actions directly, the LLM acts as a high-level optimizer over a restricted parameter set, translating natural-language guidance and recent terrain-dependent behavior into bounded edits of MPC weights and bounds, in the spirit of recent LLM-based optimization methods~\cite{Yang2023OPRO,AhmadiTeshnizi2023OptiMUS} and safety-guarded parameter tuning in autonomous systems~\cite{Baumann2025Enhancing}. The low-level neural MPC then closes the loop with hard feasibility constraints, while the language-guided regulator enables fast adaptation across terrain geometries and task preferences.

In summary, our full contributions are as follows:
\begin{itemize}[leftmargin=*]

  \item \textbf{Terrain-conditioned neural kinematics.}  
  We develop a map-attached neural kinematics model that fuses terrain and recent motion history to capture hybrid-contact effects without explicit contact-mode enumeration.
  This yields accurate yet lightweight motion prediction as the foundation for real-time planning.

  \item \textbf{Constrained neural MPC.}  
  We embed the learned kinematics in NMPC as the state-transition constraint and solve the full problem in \(100\,\mathrm{ms}\), enabling receding-horizon control with explicit state and input constraints.

  \item \textbf{Language-guided terrain adaptation.}
  We use an LLM to adapt cost weights, input bounds, and flipper PD gains from high-level guidance and recent terrain-dependent behavior, while a safety gate enforces editable-set restrictions, clipping, rate limits, smoothing, and feasibility checks.
  This improves traversal quality by \(67\%\) over PPO and enables terrain-dependent generalization without retraining the predictor or changing the solver.
\end{itemize}

\section{Related Work}
Classical planners for articulated tracked robots with flippers typically rely on geometric reasoning and pre-defined motion primitives. Early systems combined terrain scanning with semi-autonomous flipper control and safety constraints, achieving reliable but conservative behaviors under limited perception and modeling assumptions \cite{Okada2009IROS,Pecka2016IROS}. Recent geometry-based planners explicitly encode robot–terrain contact relationships to generate real-time flipper motions on rough terrain \cite{chen2023geometry}. Hybrid sampling/optimization methods further improve feasibility on discontinuous, contact-rich surfaces \cite{LiWensing2020HS_DDP,Ding2021ICRA}, yet the reliance on hand-crafted templates and mode switching restricts expressiveness and can yield sub-optimal trajectories and unnecessary energy expenditure.

An alternative line employs RL. Template-based RL selects among a small set of discrete postures \cite{Mitriakov2021Staircase}, while continuous-action policies learn end-to-end commands \cite{Pecka2018RAL,Niroui2019DeepRL,Kalashnikov2018QTOpt,Nguyen2019RLReview}. Despite promising results, RL for flipper robots often requires terrain-specific reward shaping and long training cycles; instability and divergence are common failure modes. The recent FTR-Bench study standardizes tasks and metrics for flipper-track control and highlights the sensitivity of RL performance to reward design and training stability \cite{articlerl}. Moreover, discrete posture switching introduces additional, sometimes pointless, energy consumption.

To balance optimality and data efficiency, learning-augmented MPC integrates learned models into constrained optimization. Real-time neural MPC demonstrates that neural kinematics models can be embedded in real-time control loops for agile platforms \cite{salzmann2023neural}. Tooling such as Learning for CasADi facilitates differentiable PyTorch models inside numerical optimization, enabling neural NMPC pipelines with principled constraints and warm-starts \cite{salzmann2024l4casadi,Andersson2019CasADi,verschueren2021acados,frison2020hpipm}. These advances motivate our formulation, which preserves the structure and safety of NMPC while leveraging data-driven components \cite{chee2022knode,ren2024npc,xue2024learning,zeng2024adaptive,sun2021online}.


Recent work uses language to condition robot controllers. General VLA models such as RT-2 and OpenVLA learn to map visual observations and language commands to robot actions, improving semantic generalization from large-scale pretraining~\cite{Brohan2023RT2,Kim2024OpenVLA}. However, direct action generation does not by itself impose hard state/input constraints or solve a terrain-dependent optimal control problem. For manipulation, VLMPC parses visual--language instructions and exposes them to an MPC layer that enforces goal and constraint satisfaction~\cite{zhao2024vlmpc}. On the learning side, \emph{language-to-reward} pipelines show that LLMs can synthesize rewards or curricula that steer policy training efficiently and improve transfer~\cite{Yu2023L2R,Ma2023Eureka}. LLMs have also been explored as optimizers, either by iteratively proposing candidate solutions in natural language or by formulating solver-ready optimization models~\cite{Yang2023OPRO,AhmadiTeshnizi2023OptiMUS}. Closer to our setting, Liang et al.~\cite{liang2024lmpc} couple language feedback with MPC for data-efficient adaptation, while fielded studies deploy compact, on-board LLMs to perform low-rate, safety-guarded retuning of MPC weights and constraints without disturbing the fast control loop~\cite{Baumann2025Enhancing}. Our approach follows this pattern: the LLM provides low-frequency, bounded updates to NMPC objectives/constraints, whereas the neural NMPC backbone enforces constraints and closes the high-rate loop.

\section{Methodology}

\subsection{System Overview}

\begin{figure}[t]
    \centering
    \includegraphics[width=1.0\linewidth]{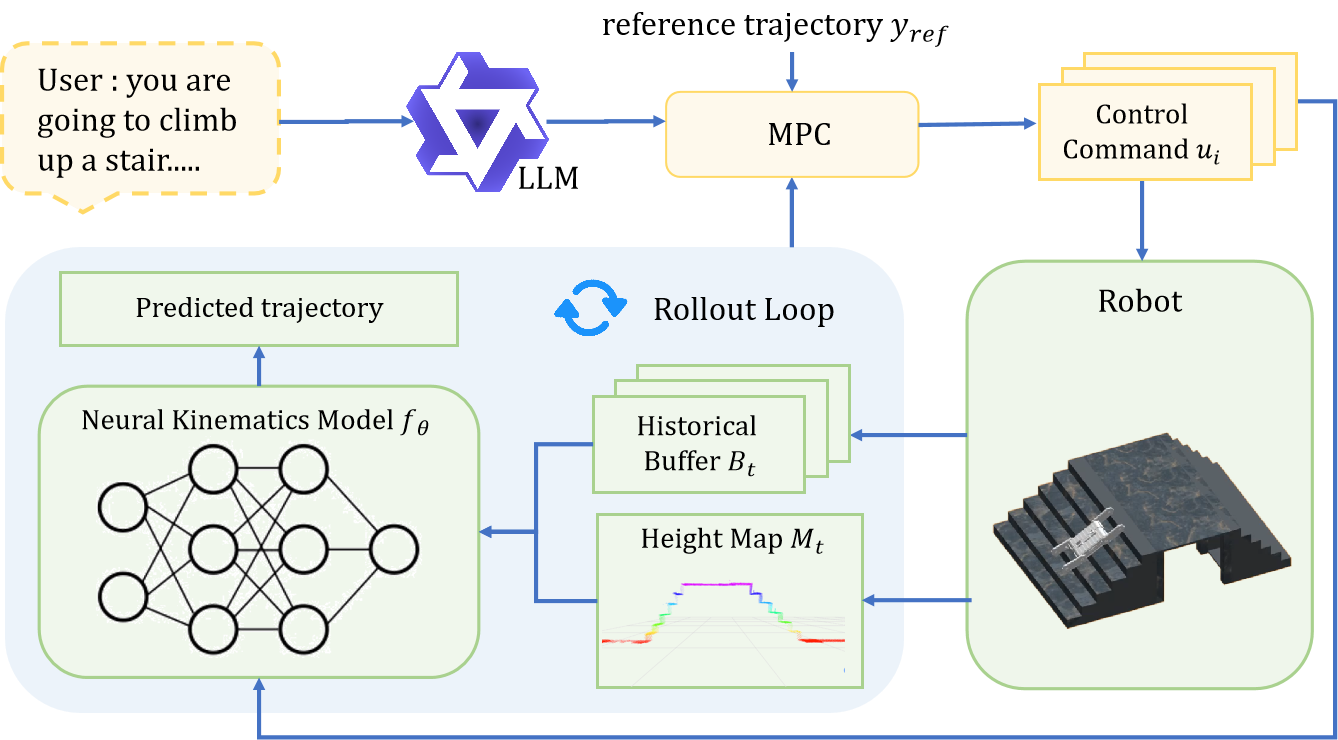}
    \caption{Overview of the proposed language-guided terrain-adaptive neural MPC system, which integrates terrain-conditioned neural kinematics, guarded parameter adaptation, and constrained motion planning.}
    \label{fig:system_overview}
\end{figure}

The proposed language-guided terrain-adaptive neural MPC architecture (Fig.~\ref{fig:system_overview}) comprises three tightly coupled layers.
First, the front end converts incoming 3D point clouds into a local height map and buffers a short history of recent robot states and control commands. 
Second, a compiled neural kinematics consumes the height map together with this history to provide short-horizon state increments in a map-attached frame (Sec.~\ref{sec:modeling}). 
Third, this kinematics is embedded as the model constraint within an NMPC, which solves a finite-horizon optimal control problem with multi-objective costs and hard feasibility constraints, applies the first control input, and then shifts the horizon as new measurements arrive (Sec.~\ref{subsec:mpc_ff}). 
An in-the-loop LLM serves as a terrain-adaptation regulator: given language guidance, a compact summary of the recent terrain-dependent state–input history, and the current parameter vector, it proposes edits to a restricted set of weights, bounds, and flipper PD gains; a guarded interface enforces value ranges, change-rate limits, and basic consistency before any update is applied (Sec.~\ref{subsec:llm_adaptation}).
The full control loop implementing these interactions is detailed in Algorithm~\ref{alg:llm_mpc}.

\subsection{Modeling}\label{sec:modeling}

\subsubsection{State Representation}
\begin{figure}[t]
    \centering
    \includegraphics[width=\linewidth,height=0.5\textheight,keepaspectratio]{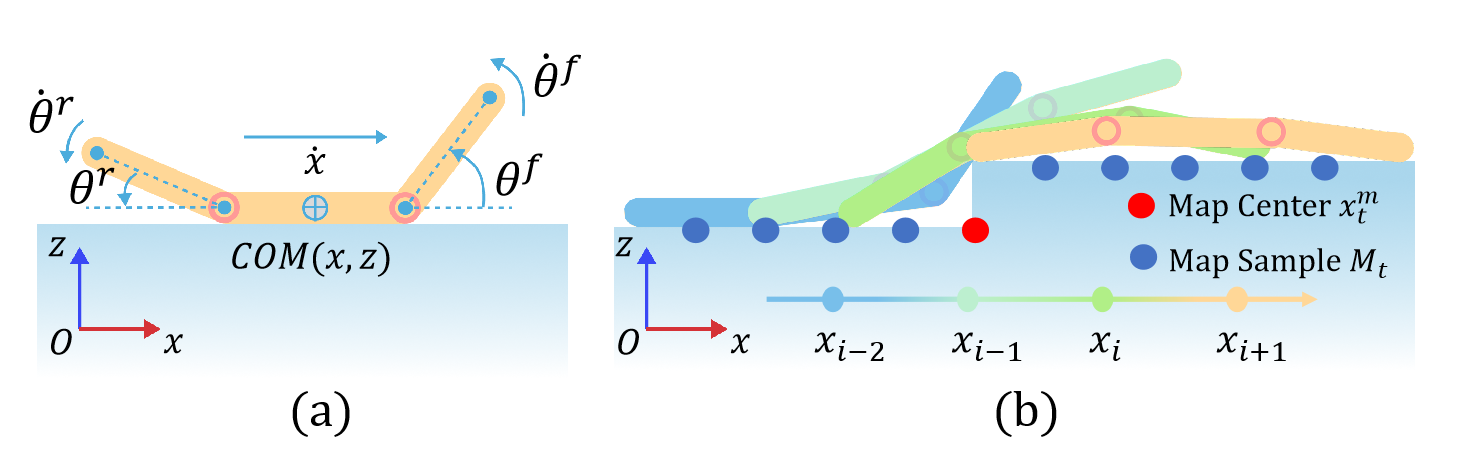}
    \caption{(a) Planar configuration of the articulated tracked robot in the sagittal $x$--$z$ plane. 
    The task variables are the COM position $(x,\,z)$, the pitch angle $\theta_{p}$, and the front and rear flipper angles $\theta^{f},\,\theta^{r}$, with motion variables $\dot{x},\,\dot{\theta}^{f},\,\dot{\theta}^{r}$. 
    (b) Map-attached representation for prediction and control. 
    At control step $t$, the elevation sequence $M_t$ is anchored at the longitudinal COM position ${x}_{t}$, yielding a \emph{fixed} terrain profile $M_{t+j}\equiv M_t$ over the horizon.}
    \label{fig:system_formulation}
\end{figure}

Explicitly first principles modeling the discontinuous contacts and support-mode switches is too complex to run in real time. We therefore adopt a lightweight \emph{neural kinematics} model conditioned on a local height sequence sampled from the global map. The state is restricted to the 2D sagittal ($x$--$z$) plane, 
which is sufficient for capturing the dominant motion of articulated tracked robots (cf. Fig.~\ref{fig:system_formulation}). This restriction is also consistent with our platform: the coaxial flipper pair cannot actively control roll, so roll stability must be handled primarily by selecting a globally roll-favorable path. Rather than addressing this global planning, our work serves as a downstream module, focusing on generating the appropriate articulated motions to track predefined trajectories and negotiate obstacles.

At each control cycle $t$, we anchor at the robot’s current longitudinal position $x_t^{m}$, 
corresponding to the $x$-coordinate of its center of mass (COM).
We take $n_m$ samples ahead and $n_m$ samples behind, spaced by $d_m$, to form an elevation sequence $M_t$ that is \emph{fixed} over the prediction horizon (H):
\begin{subequations}\label{equ:map_constant}
\begin{align}
&M_t = \left\{h(x_t^{m}+i d_m) \  | \  i=-n_m,\dots,0,\dots,n_m  \right\}, \label{equ:map_constant_a}\\
&M_{t+j} \equiv M_t,\quad j=0,\ldots,H. \label{equ:map_constant_b}
\end{align}
\end{subequations}
Here \(M_t\!\in\!\mathbb{R}^{2n_m+1}\) is the sampled height sequence
(cf.~\eqref{equ:map_constant}), where each entry is the ground height
returned by \(h(\cdot)\) at a longitudinal coordinate. To represent the robot's position relative to this terrain profile within a prediction horizon,
we define a local longitudinal coordinate
\begin{align}
\tilde{x}_{t+j} := x_{t+j} - x^{m}_t,\qquad j=0,\ldots,H.
\label{equ:rel_longitudinal_a}
\end{align} 
Based on these coordinates, we define two types of vectors:  
(1) a task vector $s_t$ that describes the robot’s absolute state, and  
(2) a state vector $\tilde{s}_t$ that describes the robot’s state relative to the sampled map, using local longitudinal offsets (anchored at \(\tilde{x}_t=0\)), 
together with their associated control inputs:
\begin{align}
\mathbf{s}_{t+j}
&=\big[x_{t+j},\,z_{t+j},\,\theta_{p,t+j},\,\theta^{f}_{t+j},\,\theta^{r}_{t+j}\big]^{\!\top},
\label{equ:task_vec}\\[3pt]
\tilde{\mathbf{s}}_{t+j}
&=\big[\tilde{x}_{t+j},\,z_{t+j},\,\theta_{p,t+j},\,\theta^{f}_{t+j},\,\theta^{r}_{t+j}\big]^{\!\top},
\label{equ:state_map}\\
\mathbf{u}_{t+j}
&=\big[v_{t+j},\,\omega^{f}_{t+j},\,\omega^{r}_{t+j}\big]^{\!\top}.
\label{equ:input_cmd}
\end{align}
for \(j=0,\ldots,H\).
To provide the learned kinematics and the NMPC with short-term motion context,
we maintain a length-$k$ buffer of recent states and inputs:
\begin{equation}
B_t := \{\tilde{s}_{t-k+1:t},\,\mathbf{u}_{t-k+1:t}\},
\label{equ:buffer_def}
\end{equation}
where \(k\) is the buffer size (number of most recent time steps).

\subsubsection{Neural Kinematics Model}
We employ a learned discrete-time kinematic model \(f_{\theta}\) that predicts the task-space increment at each stage. 
The model is queried with: (i) the local terrain sequence \(M_t\) held fixed over the prediction horizon \eqref{equ:map_constant}; 
(ii) a short history buffer \(B_t\) capturing recent motion context; and
(iii) the stage control \(\mathbf{u}_{t+j}\).
The per-stage prediction and update are
\begin{samepage}
\begin{subequations}\label{eq:nn_increment_update_t}
\begin{align}
\Delta\widehat{\mathbf{s}}_{t+j} &= f_{\theta}\!\big(M_t,\,B_t,\,\mathbf{u}_{t+j}\big),
\label{eq:nn_increment_update_t_a}\\
\widehat{\mathbf{s}}_{t+j+1} &= \mathbf{s}_{t+j} + \Delta\widehat{\mathbf{s}}_{t+j},
\qquad j=0,\ldots,H-1.
\label{eq:nn_increment_update_t_b}
\end{align}
\end{subequations}
\end{samepage}
where each \(\widehat{\mathbf{s}}_{t+j}\) has the same physical components as
\(\tilde{\mathbf{s}}_{t+j}\)
(\(\tilde{x},\,z,\,\theta_p,\,\theta^f,\,\theta^r\)).

To guide the network toward stable and control-aware predictions, 
we design a composite loss that balances accuracy, smoothness, and robustness.
The details are summarized in Table~\ref{tab:loss_components}. We use a diagonal positive weight matrix $\mathbf{W_{fit}}$.
The positive part is defined as
\begin{equation}
(a)_+ = \max(a,0).
\label{eq:pos_part}
\end{equation}

Elementwise clipping is
\begin{equation}
\mathrm{clip}(z;\delta)=\operatorname{sign}(z)\,\min(|z|,\delta).
\label{eq:clip}
\end{equation}

The short-horizon forward-motion trend is
\begin{equation}
\overline{\Delta x}_{i}=\tfrac{1}{2}\big(\Delta x_{i}+\Delta x_{i-1}\big),
\label{eq:trend}
\end{equation}
where $\Delta x_{i}$ denotes the $x$–component of $\Delta\mathbf{s}_{i}$.
\emph{Fit} ($\mathcal{L}_{\text{fit}}$) is a weighted MSE on the predicted increments, 
letting critical channels receive larger weights.  
\emph{Overshoot} ($\mathcal{L}_{\text{over}}$) penalizes only \emph{positive} 
forward-direction bias $(\Delta\hat{x}-\Delta x)_+^2$, discouraging overly aggressive advancement.  
\emph{Smoothness} ($\mathcal{L}_{\text{sm}}$) aligns the predicted forward increment 
$\Delta\hat{x}$ with the local trend $\overline{\Delta x}$ to reduce jitter across steps.  
\emph{Magnitude} ($\mathcal{L}_{\text{mag}}$) penalizes clipped predicted increments 
to keep outputs within plausible ranges and improve robustness to outliers.  
\emph{Regularization} ($\mathcal{L}_{\text{reg}}$) applies $\ell_2$ weight decay on $\theta$ to curb overfitting.

\begin{table}[b]
\centering
\caption{Control-aware training loss terms. Sums over $i$ run across the training mini-batch.}
\label{tab:loss_components}
\setlength{\tabcolsep}{24pt}
\renewcommand{\arraystretch}{1.2}
\begin{tabular}{ll}
\toprule
\textbf{Term} & \textbf{Definition} \\
\midrule
$\mathcal{L}_{\text{fit}}$  &
$\sum_{i}\|\Delta\mathbf{s}_{i+1}-\Delta\hat{\mathbf{s}}_{i+1}\|_{\mathbf{W
_{fit}}}^{2}$ \\
$\mathcal{L}_{\text{over}}$ &
$\lambda_{\text{over}}\!\sum_{i}(\Delta\hat{x}_{i+1}-\Delta x_{i+1})_{+}^{2}$ \\
$\mathcal{L}_{\text{sm}}$   &
$\lambda_{\text{sm}}\!\sum_{i}\|\Delta\hat{x}_{i+1}-\overline{\Delta x}_{i}\|_{2}^{2}$ \\
$\mathcal{L}_{\text{mag}}$  &
$\lambda_{\text{mag}}\!\sum_{i}\|\mathrm{clip}(\Delta\hat{\mathbf{s}}_{i+1};\delta)\|_{2}^{2}$ \\
$\mathcal{L}_{\text{reg}}$  &
$\lambda_{\text{reg}}\|\theta\|_{2}^{2}$ \\
\bottomrule
\end{tabular}
\end{table}

\subsection{Model Predicted Control}\label{subsec:mpc_ff}

At each control step \(t\), the NMPC optimizes over the task-space states \(\mathbf{s}_{t+j}\) 
and control inputs \(\mathbf{u}_{t+j}\) defined in Sec.~\ref{sec:modeling}.
It uses the learned kinematics 
(\ref{eq:nn_increment_update_t_a}–\ref{eq:nn_increment_update_t_b})
together with the fixed height sequence \(M_t\) (\ref{equ:map_constant})
to solve a finite-horizon problem with multi-objective costs and strict feasibility constraints.
Input penalties are centered around a PD feedforward term \(\mathbf{u}_{\mathrm{ff},t+j}\).

\begin{subequations}\label{eq:nmpc_all_in_one}
\begin{align}
\underset{\mathclap{\substack{\hat{s}_{t:t+H}\\ \mathbf{u}_{t:t+H}}}}{\min}\;
& \sum_{j=0}^{H-1}\!\Big(
      \|\hat{s}_{t+j}-\mathbf{s}^{\mathrm{ref}}_{t+j}\|_{\mathbf{W}}^{2}
    + \|\mathbf{u}_{t+j}-\mathbf{u}_{\mathrm{ff},t+j}\|_{\mathbf{R}}^{2}
  \Big) \notag\\
& \; + \|\hat{s}_{t+H}-\mathbf{s}^{\mathrm{ref}}_{t+H}\|_{\mathbf{W}_e}^{2}, \label{eq:nmpc_cost}
\end{align}

\begin{align}
 \text{s.t.}\;\; & \mathbf{B_t} = {\mathbf{B}_t^{\mathrm{ini}}}, \label{eq:nmpc_init}
\end{align}

\begin{align}
\hat{s}_{t+j+1} &= \hat{s}_{t+j} + f_\theta\!\big(M_t,\hat{B_t},\mathbf{u}_{t+j}\big), \label{eq:nmpc_dyn}\\[-2pt]
& \text{for } j=0,\ldots,H-1. \notag
\end{align}

\begin{align}
\hat{B_t} := \{\hat{s}_{t-k+1:t},\,\mathbf{u}_{t-k+1:t}\},
\label{equ:buffer_digui}
\end{align}

\begin{align}
\mathbf{u}^{\mathrm{lb}} &\le \mathbf{u}_{t+j} \le \mathbf{u}^{\mathrm{ub}}, \label{eq:nmpc_input}
\end{align}

\begin{align}
\mathbf{s}^{\mathrm{lb}} &\le \hat{s}_{t+j} \le \mathbf{s}^{\mathrm{ub}}. \label{eq:nmpc_state}
\end{align}
\end{subequations}

\subsubsection{Cost and Reference}
We use diagonal state weights $\mathbf{W}$ (terminal $\mathbf{W}_e$) on the five task variables and diagonal input penalties $\mathbf{R}$. The input term is centered at a feedforward sequence $\mathbf{u}_{\mathrm{ff},t+j}=[\,u^{\mathrm{ff}}_{f,t+j},\,u^{\mathrm{ff}}_{r,t+j}\,]^\top$ aligned to the NMPC grid. For the flippers, $\mathbf{u}_{\mathrm{ff}}$ is generated by per-joint PD laws:
\begin{subequations}\label{eq:pd_ff}
\begin{align}
u^{\mathrm{ff}}_{f,t+j}
&= K^{f}_{P}\big(\theta^{f,\mathrm{ref}}_{t+j}-\theta^{f}_{t}\big)
 + K^{f}_{D}\big(\omega^{f,\mathrm{ref}}_{t+j}-\omega^{f}_{t}\big),
\label{eq:pd_ff_a}\\
u^{\mathrm{ff}}_{r,t+j}
&= K^{r}_{P}\big(\theta^{r,\mathrm{ref}}_{t+j}-\theta^{r}_{t}\big)
 + K^{r}_{D}\big(\omega^{r,\mathrm{ref}}_{t+j}-\omega^{r}_{t}\big).
\label{eq:pd_ff_b}
\end{align}
\end{subequations}
Here $(\theta^{f}_{t},\omega^{f}_{t})$ and $(\theta^{r}_{t},\omega^{r}_{t})$ are the measured joint positions and velocities at time $t$, while $(\theta^{f,\mathrm{ref}}_{t+j},\omega^{f,\mathrm{ref}}_{t+j})$ and $(\theta^{r,\mathrm{ref}}_{t+j},\omega^{r,\mathrm{ref}}_{t+j})$ are the stage-$t{+}j$ references. To suppress first-stage jerk on the chassis, the forward-velocity reference is ramped within the horizon:
\begin{equation}
v^{\mathrm{ref}}_{t+j} \;=\; v_0 + \frac{j}{H}\big(v_{\mathrm{ref}}-v_0\big), \qquad j=0,\ldots,H-1.
\end{equation}
with $v_0$ the measured speed at time $t$. Before entering the cost, each $\mathbf{u}_{\mathrm{ff},t+j}$ is clipped to the actuator box in~\eqref{eq:nmpc_input} and subjected to the same rate limit $\Delta u_{\max}$ (with smoothing factor $\alpha_{\mathrm{sm}}$) as the applied control, which keeps the penalty center feasible and reduces early-stage optimizer effort.

\subsubsection{Constraints}
We enforce feasibility by clamping the initial state to the current measurement $\hat{\mathbf{s}}_t$ and constraining all control inputs within a bounded feasible set. At each stage,
\(\mathbf{u}_{t+j} = [\,v_{t+j},\,\omega^{f}_{t+j},\,\omega^{r}_{t+j}\,]^\top\)
must lie in
\(\mathcal{U} = \{\mathbf{u}\mid \mathbf{u}^{\mathrm{lb}} \le \mathbf{u} \le \mathbf{u}^{\mathrm{ub}}\}\),
which includes explicit forward-speed bounds
\(v_{\min} \le v_{t+j} \le v_{\max}\).
These bounds are part of the editable parameter set (Table~\ref{tab:llm_params}) and can be adjusted online under safety gating.

\subsection{Language-Guided Terrain Adaptation via LLM}
\label{subsec:llm_adaptation}

\begin{figure}[t]
    \centering
    \includegraphics[width=\linewidth,height=0.5\textheight,keepaspectratio]{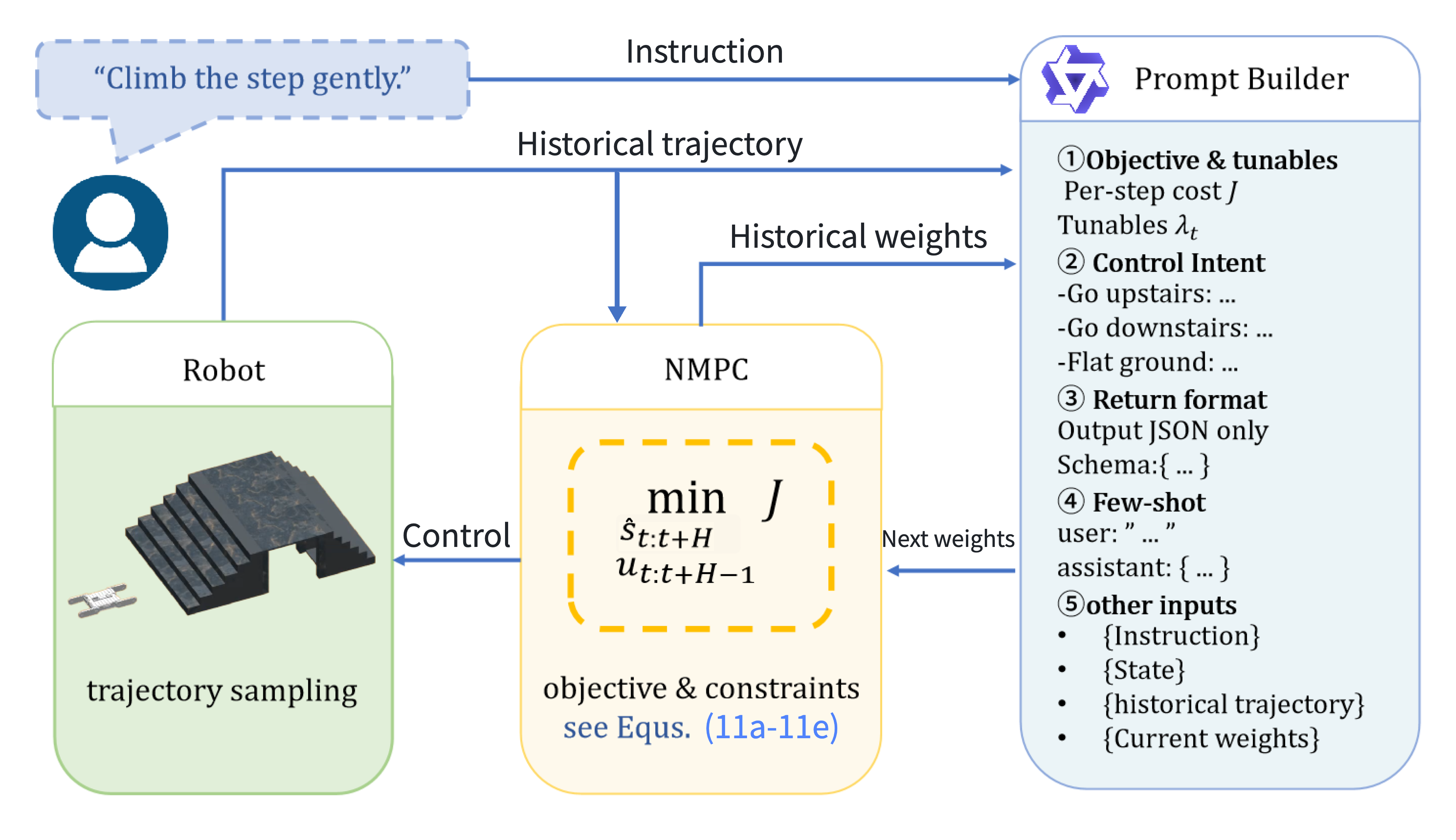}
    \caption{Language guidance and recent terrain-dependent history are formatted by the Prompt Builder and sent to the LLM. The LLM returns a JSON of neural MPC parameter updates (tracking weights, input penalties, flipper PD gains, smoothing/rate limits, and speed reference/bounds), which overwrite the current set. Neural MPC applies these updates only to costs and constraints (see Eqs.~\ref{eq:nmpc_cost}--\ref{eq:nmpc_state}), solves the horizon, and sends control commands; parameters persist until new guidance arrives.}
    \label{fig:llm_adaptation}
\end{figure}

Although neural MPC provides a principled way to enforce constraints and balance multiple objectives, its performance on ATRs is sensitive to cost weights and constraint bounds. This sensitivity becomes more severe under out-of-distribution terrain, where a fixed objective may no longer express the desired trade-off among speed, stability, and impact. Instead of using an LLM as an open-loop action policy, we use it as a structured terrain-adaptation assistant: following the idea that LLMs can propose and refine optimization variables or solver-ready formulations~\cite{Yang2023OPRO,AhmadiTeshnizi2023OptiMUS}, the model only suggests parameter edits for an existing constrained neural MPC controller. This yields a rapid and interpretable language-guided adaptation layer without altering the MPC formulation or bypassing its feasibility constraints (Fig.~\ref{fig:llm_adaptation}).

When a new operator instruction arrives, the controller assembles a compact prompt \(Q_t\) from: (i) the current measurement \((\tilde{\mathbf{s}}_{t},\,\mathbf{u}_{t})\); (ii) the length-\(k\) history buffer \(B_t\); (iii) the active parameter vector \(\boldsymbol{\lambda}_{t-1}\) that instantiates the NMPC weights \((\mathbf{W},\mathbf{W}_e,\mathbf{R})\) and bounds \((v_{\min}, v_{\max})\); and (iv) a concise parameter dictionary and optimization template (objective and constraints in~\eqref{eq:nmpc_cost}--\eqref{eq:nmpc_state}), followed by the operator instruction \(\mathrm{instr}_t\). The prompt therefore asks the LLM to solve a small, semantic parameter-selection problem rather than the continuous NMPC itself. The LLM is treated as a conditional generator of parameter updates; its output is restricted to a predefined subset and passed through a safety gate that enforces bounds, per-update rate limits, and smoothing before any change takes effect.

\begin{table}[b]
\centering
\caption{Parameters adapted online by the LLM.}
\label{tab:llm_params}
\setlength{\tabcolsep}{6pt}
\renewcommand{\arraystretch}{1.15}
\small
\begin{tabular}{p{0.40\linewidth} p{0.50\linewidth}}
\toprule
\textbf{Category} & \textbf{Parameters} \\
\midrule
State tracking weights &
$W_{\mathrm{fwd}},\, W_{z},\, W_{\theta_p},\, W_{\theta_f},\, W_{\theta_r}$ \\
Control penalties &
$R_{v},\, R_{\omega^f},\, R_{\omega^r}$ \\
Feedforward shaping &
$K^{f}_{P},\, K^{f}_{D},\, K^{r}_{P},\, K^{r}_{D}$ \\
Smoothing and limits &
$\Delta u_{\max},\, \alpha_{\mathrm{sm}}$ \\
Velocity reference/bounds &
$v_{\mathrm{ref}},\, v_{\min},\, v_{\max}$ \\
\bottomrule
\end{tabular}
\end{table}

\paragraph*{Prompt construction and parameter extraction}
Using the above quantities, prompt construction and parameter extraction are written as
\begin{subequations}\label{eq:llm_io_chain}
\begin{align}
Q_t &= \mathrm{PromptBuild}\!\big(B_t,\,M_t,\,\boldsymbol{\lambda}_{t-1},\,\mathrm{instr}_t), \label{eq:prompt_build}\\
\tilde{\boldsymbol{\lambda}}_t &= \mathrm{ParamExtract}\!\big(\mathrm{LLM}(Q_t)\big)\big|_{\mathcal{S}}. \label{eq:param_extract}
\end{align}
\end{subequations}
where \(\mathcal{S}\) is the editable set in Table~\ref{tab:llm_params} (state-tracking weights, input penalties, flipper PD gains, smoothing factors, and velocity bounds). All other components remain fixed.

\paragraph*{Guarded update}
The suggested edits are passed through a safety gate that enforces validity, magnitude limits, and temporal smoothness:
\begin{subequations}\label{eq:llm_guard_min}
\begin{align}
\bar{\boldsymbol{\lambda}}_t
&= \mathrm{clip}\!\left(\tilde{\boldsymbol{\lambda}}_t;\ \boldsymbol{\lambda}^{\mathrm{lb}},\,\boldsymbol{\lambda}^{\mathrm{ub}}\right), \label{eq:clip}\\
\hat{\boldsymbol{\lambda}}_t
&= \mathrm{proj}_{\|\cdot-\boldsymbol{\lambda}_{t-1}\|_\infty \le \Delta u_{\max}}\!\left(\bar{\boldsymbol{\lambda}}_t\right), \label{eq:rate}\\
\boldsymbol{\lambda}_t
&= (1-\alpha_{\mathrm{sm}})\,\boldsymbol{\lambda}_{t-1} + \alpha_{\mathrm{sm}}\,\hat{\boldsymbol{\lambda}}_t. \label{eq:smooth}
\end{align}
\end{subequations}
with elementwise bounds \((\boldsymbol{\lambda}^{\mathrm{lb}},\boldsymbol{\lambda}^{\mathrm{ub}})\), a per-update \(\ell_\infty\) cap \(\Delta u_{\max}\), and smoothing factor \(\alpha_{\mathrm{sm}}\!\in\!(0,1]\). Feasibility checks (e.g., \(v_{\min}<v_{\max}\) and \(\mathbf{W},\mathbf{W}_e,\mathbf{R}\succ0\)) are verified before applying~\eqref{eq:smooth}. Only \(\boldsymbol{\lambda}\) is updated; the perception stack, the learned kinematic model \(f_\theta\), and the NMPC solver remain unchanged. This mechanism enables fast, language-guided terrain adaptation while preserving the underlying control pipeline.

\begin{algorithm}[t]
\caption{Language-guided terrain-adaptive neural MPC pipeline}\label{alg:llm_mpc}
\begin{algorithmic}[1]
\State \textbf{Offline:} \(M \leftarrow \mathrm{ExtractHeightMap}(\mathcal{P})\);
       \(\{\mathbf{s}^{\mathrm{ref}}_{t+j}\}_{j=0}^{H} \leftarrow \mathrm{GenerateReference}(M)\) \label{step:offline}
\State \textbf{Init:} buffer \(B_t\), model \(f_\theta\), params \(\boldsymbol{\lambda}_0\), horizon \(H\) \label{step:init}
\While{robot active} \label{step:while}
  \State Update \(B_t\) and local map slice \(M_t\) \label{step:updateBt}
  \If{new instruction} \label{step:if_instr}
    \State \(Q_t \leftarrow \mathrm{BuildPrompt}(\mathrm{instr}_t, M_t, B_t, \boldsymbol{\lambda}_{t-1})\) \label{step:prompt}
    \State apply guarded update (\ref{eq:clip}–\ref{eq:smooth}) \label{step:update_param}
  \EndIf
  \State Build PD feedforward and ramp \(v^{\mathrm{ref}}\) \label{step:pdff}
  \For{$j=0$ to $H{-}1$} \label{step:for}
    \State \(\widehat{\mathbf{s}}_{t+j+1} \gets \widehat{\mathbf{s}}_{t+j} + f_\theta(M_t, \widehat{B}_t, \mathbf{u}_{t+j})\)
    \State \(\widehat{B}_t \gets \{\widehat{\mathbf{s}}_{t-k+1:t},\,\mathbf{u}_{t-k+1:t}\}\) \label{step:updatehatBt}
  \EndFor
  \State Solve NMPC (\ref{eq:nmpc_cost}–\ref{eq:nmpc_state}); apply smoothing; send \(\mathbf{u}_t\) \label{step:solve_nmpc}
\EndWhile
\end{algorithmic}
\end{algorithm}

\noindent\textbf{Runtime loop.}
Algorithm~\ref{alg:llm_mpc} summarizes the closed-loop procedure.
Offline, a local height map and reference profiles are prepared (Step~\ref{step:offline}–\ref{step:init}).
Online, each cycle updates the history buffer \(B_t\) and the fixed terrain slice \(M_t\) (Step~\ref{step:updateBt}).
If an operator instruction arrives, the controller freezes actuation, builds a prompt \(Q_t\) (Step~\ref{step:prompt}),
parses the LLM output and applies the guarded update (Step~\ref{step:update_param}).
It then constructs the flipper PD feedforward \(\mathbf{u}_{\mathrm{ff}}\) (Step~\ref{step:pdff}),
rolls the learned kinematics over the horizon (Step~\ref{step:for}–\ref{step:updatehatBt}),
and solves the NMPC (Step~\ref{step:solve_nmpc}).


\section{Experiments and Results}

\subsection{Experimental Setup}\label{sec:exp_setup}

\subsubsection{Platform and Simulation Setup}
We use the tracked rescue platform from~\cite{NuBot_Rescue_Gazebo} (Fig.~\ref{fig:robot_overview}), featuring four independently actuated flippers and a symmetric chassis. 
Simulation scenes follow~\cite{chen2023geometry}; the robot traverses terrains represented as height maps 
with a sampling interval \(d_m = 0.0625\) m \eqref{equ:map_constant_a} and the full NMPC loop runs within \(100\)~ms per cycle.

\subsubsection{Hardware and Software}
All components run on \emph{ROS Noetic}, Ubuntu~20.04. 
The NMPC is modeled in \texttt{CasADi}~\cite{Andersson2019CasADi}; the learned kinematic model is compiled to solver-compatible code via \texttt{L4CasADi}~\cite{salzmann2023neural,salzmann2024l4casadi}; and the resulting nonlinear program is solved online with \texttt{ACADOS}~\cite{verschueren2021acados}. 
Experiments run on an Intel i5 CPU with an NVIDIA RTX~2080Ti GPU.

\subsubsection{Default Parameters}
Unless otherwise stated, we use horizon \(H{=}12\) with \(\Delta t{=}0.1\)~s; state–tracking weights 
\(W_{\mathrm{fwd}}{=}3.0,\; W_{z}{=}0.5,\; W_{\theta_p}{=}15.0,\; W_{\theta_f}{=}5.0,\; W_{\theta_r}{=}5.0\); 
input penalties \(R_{v}{=}10^{-3},\; R_{\omega^f}{=}2{\times}10^{-2},\; R_{\omega^r}{=}2{\times}10^{-2}\); 
flipper PD gains \(K^{f}_{P}{=}K^{r}_{P}{=}2.0,\; K^{f}_{D}{=}K^{r}_{D}{=}0.3\); 
rate/smoothing \(\Delta u_{\max}{=}0.18,\; \alpha_{\mathrm{sm}}{=}0.8\); 
and speed targets \(v_{\mathrm{ref}}{=}0.30\)~m/s with initial command \(v_0{=}0.12\)~m/s. 
These categories correspond to the tunable entries in Table~\ref{tab:llm_params}.

\subsubsection{Language Model}
Language-guided terrain adaptation is realized by the \emph{DashScope Qwen-Plus} LLM~\cite{qwen}, which generates bounded updates to neural MPC weights and soft constraints from sliding-window summaries of terrain-dependent behavior and high-level task instructions.

\subsection{Data Collection and Training Details}

We train a next-step kinematic predictor as supervised regression that maps a history-augmented input to the 5-dimensional task state at \(t{+}1\). The input concatenates a 64-bin local height sequence, recent longitudinal positions, recent commands, and short-horizon robot states (Sec.~\ref{sec:modeling}); samples are recorded at 10 Hz and post-processed into the compact representation.  

The dataset comprises 5k simulated samples from stair ascent/descent, split 95:5 for training/testing. The predictor is a feedforward MLP with residual connections and a linear 5D output head, trained with Adam (lr $1\times10^{-4}$, batch 2048) using the control-aware loss in Table~\ref{tab:loss_components}; generalization is evaluated separately in Sec.~\ref{sec:generative}.

\subsection{Evaluation and Analysis of Neural Kinematics}
\label{sec:model_analysis}

To isolate the effect of the loss (symbols as in Table~\ref{tab:loss_components}), we compare three configurations:
\emph{Type~1} ($\mathcal L_{\text{fit}}$ only), \emph{Type~2} ($\mathcal L_{\text{fit}}+\mathcal L_{\text{over}}+\mathcal L_{\text{mag}}+\mathcal L_{\text{reg}}$), and
\emph{Type~3} (Type~2 $+\mathcal L_{\text{sm}}$). Here $\mathcal L_{\text{fit}}$ is the fitting term, $\mathcal L_{\text{over}}$ penalizes forward overshoot, $\mathcal L_{\text{mag}}$ constrains prediction magnitude, $\mathcal L_{\text{reg}}$ is the $L_2$ regularizer, and $\mathcal L_{\text{sm}}$ encourages temporal smoothness. All models are trained and evaluated under identical data and rollout settings.

\subsubsection{Evaluation Protocol and RMSE Aggregation}
Let the prediction horizon be \(H{=}10\). For each evaluation trajectory of length \(T\),
we split it into \(S=\lfloor T/H \rfloor\) non-overlapping windows of length \(H\).
At the start of each window we freeze the height sequence and history buffer, run the model
open-loop for \(j=1,\ldots,H\), and measure step-ahead errors on the task-vector components
\(c\in\{x,z,\theta_p,\theta_f,\theta_r\}\).
For each step \(j\), we compute the RMSE across the \(S\) windows between the \(j\)-step
predictions and ground truth for each component. We compare the per-step RMSE across
the $x$ direction, the $z$ direction, the pitch angle, and the front/rear flipper angles.

\begin{figure}[t]
\centering
\includegraphics[width=.45\textwidth]{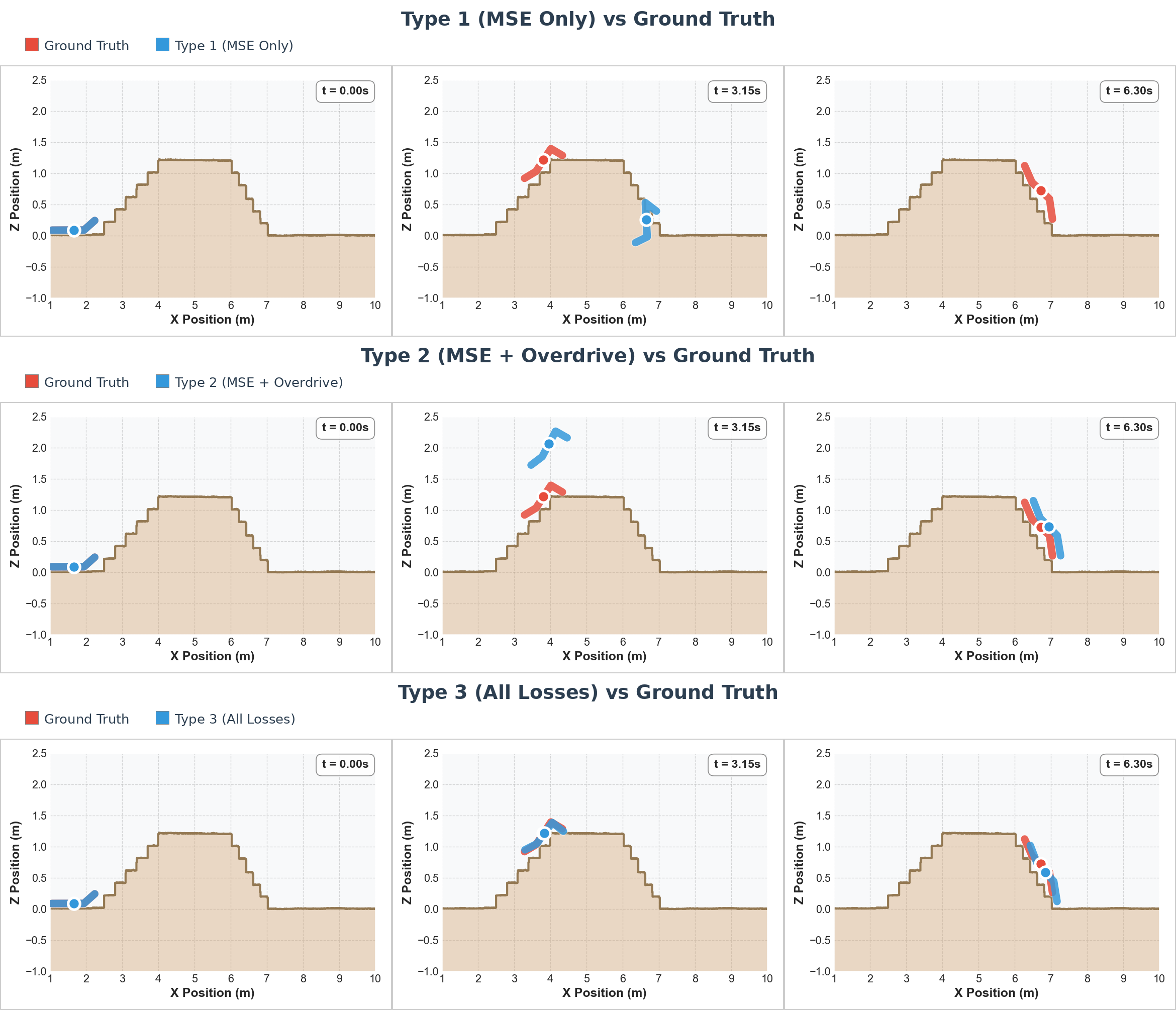}
\caption{Visualizing prediction vs.\ ground truth: from one continuous prediction rollout,
we randomly sample three discrete time steps and show the predicted states (blue) 
against the ground-truth trajectory (red) for each loss design (Type 1/2/3).
}
\label{fig:trajectory_3x3}
\end{figure}

\begin{figure}[t]
    \centering
    \includegraphics[width=1.0\linewidth]{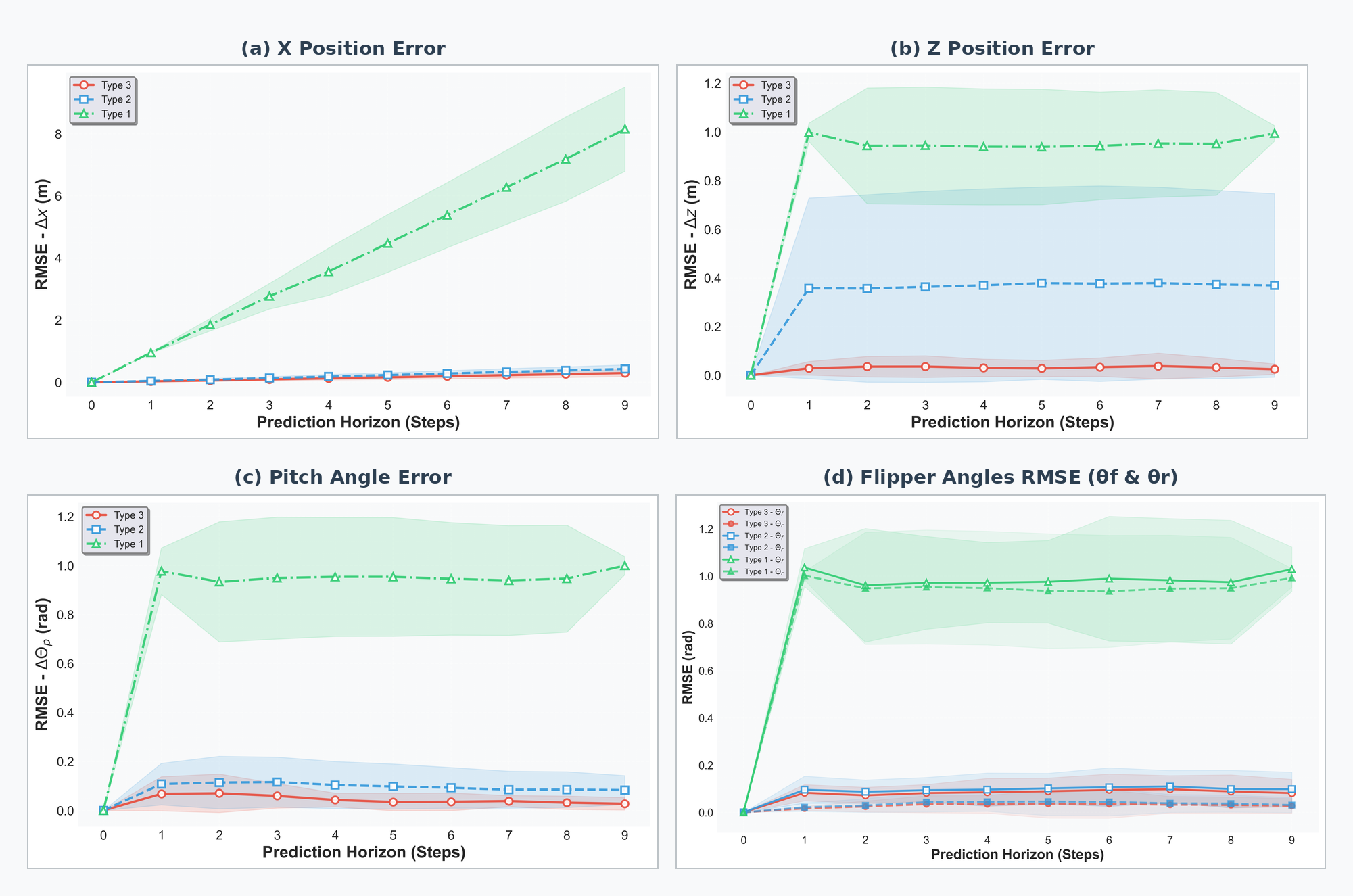}
    \caption{
        Per-step RMSE ($j{=}1\ldots10$) under three loss designs: Type~1 ($\mathcal L_{\text{fit}}$),
        Type~2 (Type~1 $+\mathcal L_{\text{over}}+\mathcal L_{\text{mag}}+\mathcal L_{\text{reg}}$),
        and Type~3 (Type~2 $+\mathcal L_{\text{sm}}$). Subplots show (a) $x$-direction error,
        (b) $z$-direction error, (c) pitch-angle error, and (d) front/rear flipper angle errors.
        Curves show RMSE $\pm$ standard deviation.
    }
    \label{fig:rmse_comparison}
\end{figure}

\subsubsection{Results and Analysis}
As shown in Fig.~\ref{fig:trajectory_3x3} and Fig.~\ref{fig:rmse_comparison}, Type~1 exhibits horizon-wise drift, Type~2 reduces this drift but still shows noticeable errors, and the full loss (Type~3) yields the most accurate and stable rollouts across all state components. Meanwhile, the trajectory visualization shows the same trend: Type~1/2 diverge from the ground truth as the horizon unfolds, whereas Type~3 stays closely aligned throughout the rollout, supporting the effectiveness of our final loss design. The RMSE curves and the visualization are consistent across all variables, indicating that the errors are not dominated by a single state.

\begin{figure}[t]
    \centering
    \includegraphics[width=\linewidth]{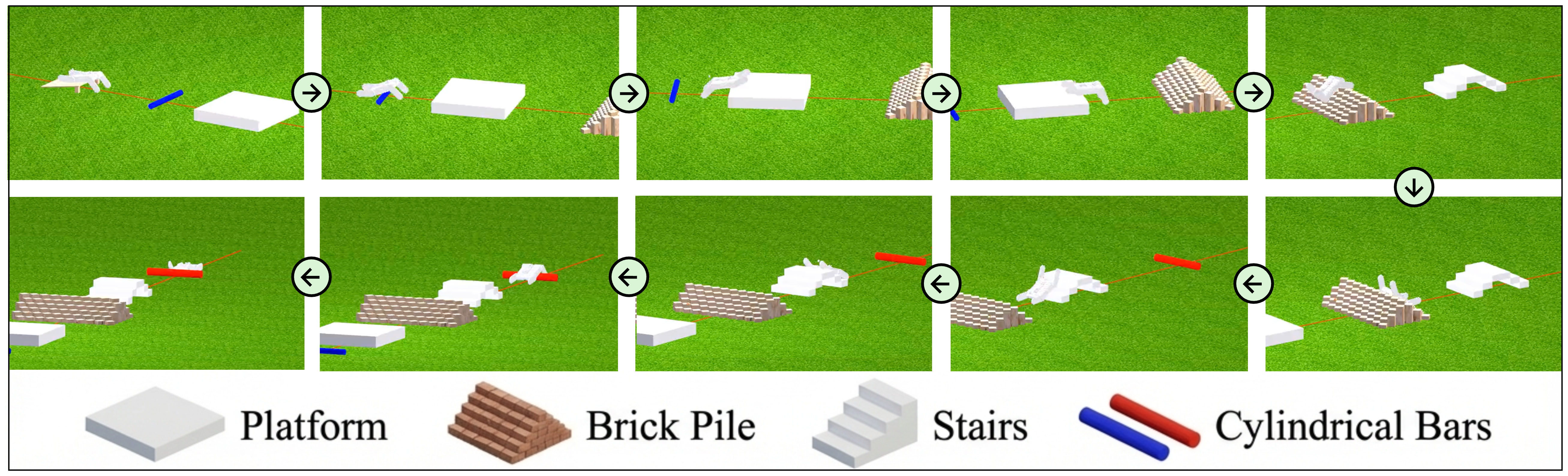}
    \caption{Generalization experiment in a single environment with multiple obstacle heights. The controller maintains feasible configurations and consistent traversal behaviors across height variations.}
    \label{fig:generative}
\end{figure}

\subsection{Comparative Experiment}
\label{sec:comparative_experiment}

We compare against a reinforcement learning baseline implemented with \emph{Proximal Policy Optimization (PPO)}~\cite{articlerl}.

\subsubsection{Evaluation Tasks}
We evaluate three scenarios: single-block crossing, stair ascent, and stair descent. Each scenario is executed three times per controller, with success defined as reaching the goal within $60$\,s without tipping.

\subsubsection{Performance Metrics}
\label{subsubsec:metrics}

To evaluate traversal quality, we record the following six raw indicators over each run:
\(\phi,\theta_p\) are the roll and pitch angles (in rad),
\(\dot{\phi},\dot{\theta_p}\) their angular velocities (in rad/s),
and \(a^{x},a^{z}\) are the linear accelerations of the robot center of mass along the forward and vertical axes (in \(\mathrm{m/s^2}\)).
For clarity, we group them into three paired terms that summarize stability, smoothness, and impact:

\begin{subequations}\label{eq:metrics}
\begin{align}
\small\emph{Stability} &= \frac{1}{2}\big(\mathrm{norm}(|\phi|_{\max}) + \mathrm{norm}(|\theta_p|_{\max})\big), \label{eq:metrics_stas}\\[2pt]
\small\emph{Swing} &= \frac{1}{2}\big(\mathrm{norm}(|\dot{\phi}|_{\max}) + \mathrm{norm}(|\dot{\theta}_p|_{\max})\big), \label{eq:metrics_swis}\\[2pt]
\small\emph{\hspace{-3em}Shock} &= \frac{1}{2}\big(\mathrm{norm}(|a^{x}|_{\max}) + \mathrm{norm}(|a^{z}|_{\max})\big).\label{eq:metrics_shos}\end{align}
\end{subequations}

Each pair of indicators is normalized to $[0,1]$ and combined into three dimensionless penalty terms (lower is better). These penalty terms are broadly applicable to most prior studies on ATRs traversal, as they capture stability, smoothness, and impact in a task-agnostic manner. All experiments are repeated three times per scenario, and the reported values are the means across repetitions.

\subsubsection{Results}
Table~\ref{tab:comparison_metrics} reports the three normalized penalties and their aggregate mean across terrains. Terrain-adaptive neural MPC improves the overall mean by \textbf{42\%} over fixed-parameter neural MPC and by \textbf{67\%} over PPO, including gains of \textbf{33\%} on ascent and \textbf{71\%} on descent over fixed-parameter neural MPC. Although the fixed controller has a slightly better single-block mean, adaptation reduces its Shock penalty from $3.0{\times}10^{-3}$ to $1.4{\times}10^{-4}$. The largest benefits therefore occur on the more dynamic stair scenarios.

\subsection{Generative Experiment}
\label{sec:generative}

We further test multiple obstacle heights in a single environment (Fig.~\ref{fig:generative}). Without task-specific retraining and using the same initial parameters, the controller maintains feasible configurations and bounded chassis pitch by coordinating front-flipper probing and rear-flipper bracing. Stable bounded LLM updates across the height variations indicate transfer to unseen geometric conditions.

\begin{table}[t]
\centering
\caption{Comparative results across terrains.}
\label{tab:comparison_metrics}
\renewcommand{\arraystretch}{1.15}
\setlength{\tabcolsep}{6pt}
\resizebox{\linewidth}{!}{%
\begin{tabular}{|l|l|c|c|c|}
\hline
\textbf{Terrain} & \textbf{Method} & \textbf{Stability} & \textbf{Swing} & \textbf{Shock} \\
\hline
\multirow{3}{*}{Single-Block}
& Terrain-adaptive neural MPC (ours) & \textbf{0.39} & 0.63 & $\mathbf{1.40{\times}10^{-4}}$ \\
\cline{2-5}
& Neural MPC (ours) & \underline{0.50 }& \textbf{0.33} & $\underline{3.04{\times}10^{-3}}$ \\
\cline{2-5}
& PPO        & 0.50 & 0.50 & 1.00 \\
\hline
\multirow{3}{*}{Stair Ascent}
& Terrain-adaptive neural MPC (ours) & \textbf{0.37} & \underline{0.08} & $\underline{2.96{\times}10^{-4}}$ \\
\cline{2-5}
& Neural MPC (ours) & 0.66 & \textbf{0.00} & \textbf{0.00} \\
\cline{2-5}
& PPO        & 0.50 & 1.00 & 1.00 \\
\hline
\multirow{3}{*}{Stair Descent}
& Terrain-adaptive neural MPC (ours) & \textbf{0.49} & \textbf{0.12} & \textbf{0.00} \\
\cline{2-5}
& Neural MPC (ours) & 0.56 & 0.98 & \underline{0.57} \\
\cline{2-5}
& PPO        & 0.50 & 0.50 & 0.76 \\
\hline
\end{tabular}
}%
\vspace{2pt}
\footnotesize\emph{Note:}  Each column corresponds to one penalty term defined in~\eqref{eq:metrics}. \textbf{bold} denotes the best result and \underline{underlined} denotes the second best.
\end{table}

\subsection{Real-Robot Validation}
\label{sec:real_robot}

\begin{figure*}[t]
    \centering
    \includegraphics[width=0.95\textwidth]{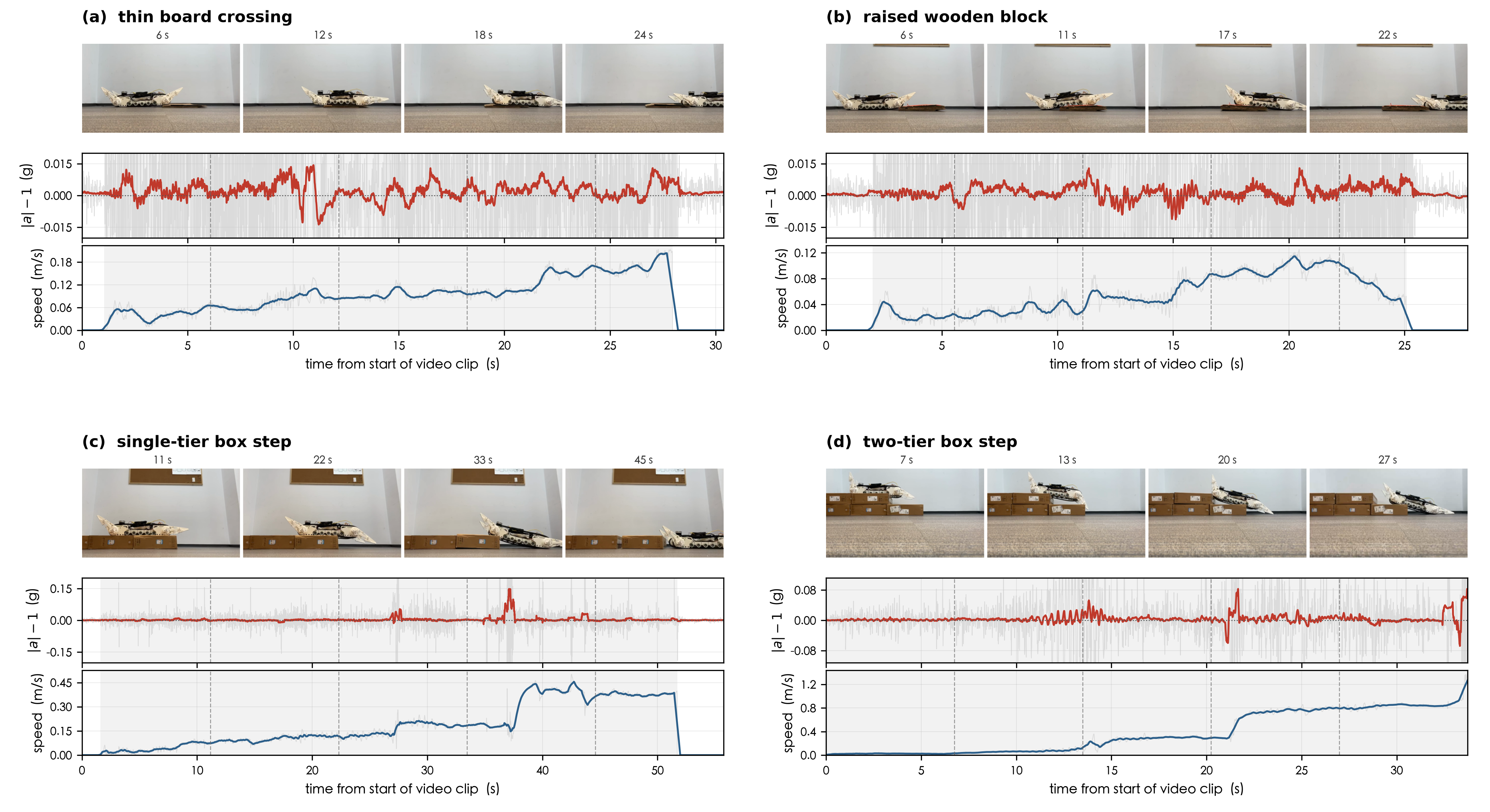}
    \caption{Real-robot traversal over four indoor obstacles. Red: Ouster-IMU acceleration-magnitude residual; blue: IMU-integrated speed magnitude used as a qualitative motion indicator.}
    \label{fig:real_robot}
\end{figure*}

We validate on the physical ATR over four indoor obstacles (Fig.~\ref{fig:real_robot}). Both signals come from the Ouster lidar IMU: the upper plots show the raw and 0.5~s-smoothed acceleration-magnitude residual $\|\bm a\|_2-1$, which indicates vibration and contact impacts, while the lower plots show an attitude-compensated, IMU-integrated three-dimensional speed magnitude. Because the latter can drift, it is used only as a qualitative motion indicator. The trials demonstrate contact-rich traversal across different obstacle geometries.

\section{Conclusion}
We proposed a language-guided terrain-adaptive neural MPC framework that brings the strengths of constrained predictive control to ATR traversal while addressing practical modeling and terrain-dependent tuning challenges. By combining terrain-conditioned neural kinematics with a language-guided LLM regulator, our method reduces the burden of scenario-specific parameter tuning while preserving feasibility and real-time performance. Furthermore, compiling the neural kinematics into the solver accelerates optimization, enabling real-time control on ATRs operating in structured, contact-rich environments. Our experiments indicate that the learned kinematics and composite loss improve multi-step prediction quality, the loss design stabilizes rollouts by curbing overshoot and encouraging smoothness, and language-guided terrain adaptation yields more stable traversal across varied terrains. Real-robot trials on four indoor obstacles further demonstrate contact-rich physical traversal across different obstacle geometries.

In future work, we plan to extend the terrain-adaptive neural MPC framework toward handling deformable or variable-geometry objects and incorporating energy-aware objectives to further improve efficiency in long-duration missions. We also aim to validate the framework on additional real-world platforms and expand the range of terrains and contact conditions.


\bibliographystyle{IEEEtran}
\bibliography{IEEEabrv,bib/bibliography}
\end{document}